\documentclass[runningheads]{llncs}
\usepackage[T1]{fontenc}
\usepackage{graphicx}
\usepackage{xcolor}
\usepackage{url}
\usepackage{wrapfig}
\usepackage{caption}
\usepackage{subcaption}

\begin{document}
\title{How Useful are Educational Questions Generated by Large Language Models?}
\titlerunning{How Useful are Educational Questions Generated by LLMs?}
\author{Sabina Elkins\inst{1, 2} \and Ekaterina Kochmar \inst{2, 3} \and \\  Iulian Serban \inst{2} \and Jackie C.K. Cheung \inst{1, 4}}
\authorrunning{S. Elkins et al.}
\institute{McGill University \& MILA (Quebec Artificial Intelligence Institute)
\and Korbit Technologies Inc. \and MBZUAI \and Canada CIFAR AI Chair}

\maketitle 
\begin{abstract} \vspace{-6mm}
Controllable text generation (CTG) by large language models has a huge potential to transform education for teachers and students alike.
Specifically, high quality and diverse question generation can dramatically reduce the load on teachers and improve the quality of their educational content.
Recent work in this domain has made progress with generation, but fails to show that real teachers judge the generated questions as sufficiently useful for the classroom setting; or if instead the questions have errors and/or pedagogically unhelpful content.
We conduct a human evaluation with teachers to assess the quality and usefulness of outputs from combining CTG and question taxonomies (Bloom's and a difficulty taxonomy).
The results demonstrate that the questions generated are high quality and sufficiently useful, showing their promise for widespread use in the classroom setting.

\keywords{Controllable Text Generation \and Personalized Learning \and Prompting \and Question Generation}
\end{abstract}

\vspace{-6mm} \section{Introduction}
The rapidly growing popularity of large language models (LLMs) has taken the AI community and general public by storm. This attention can lead people to believe LLMs are the right solution for every problem. In reality, the question of the usefulness of LLMs and how to adapt them to real-life tasks is an open one.

Recent advancements in LLMs have raised questions about their impact on education, including promising use cases \cite{baidoo_2023,terwiesch_2023,wang_2022,wang_valdez_2022}. 
A robust question generation (QG) system has the potential to empower teachers by decreasing their cognitive load while creating teaching material. 
It could allow them to easily generate personalized content to fill the needs of different students by adapting questions to Bloom's taxonomy levels (i.e., learning goals) or difficulty levels. Already, interested teachers report huge efficiency increases using LLMs to generate questions \cite{baidoo_2023,terwiesch_2023}.
These improvements hinge on the assumption that the candidates are high quality and are actually judged to be useful by teachers generally.
To the best of our knowledge, there has yet to be a study assessing how a larger group of teachers perceive a set of candidates from LLMs.\footnote{\cite{wang_valdez_2022} show that subject matter experts can't distinguish between machine and human written questions, but state that a future direction is to assess CTG with teachers.}
We investigate if LLMs can generate different types of questions from a given context that teachers think are appropriate for use in the classroom. Our experiment shows this is the case, with high quality and usefulness ratings across two domains and 9 question types.

\vspace{-2mm} \section{Background Research}
Auto-regressive LLMs are deep learning models trained on huge corpora of data. Their training goal is to predict the next word in a sequence, given all of the previous words \cite{zhang_2022}.
An example of an auto-regressive LLM is the GPT family of models, such as GPT-3. Recently, GPT-3 has been fine-tuned with reinforcement learning to create a powerful LLM called InstructGPT, which outperforms its predecessors in the GPT family \cite{ouyang_2022}. Using human-annotated data, the creators of InstructGPT use supervised learning to train a reward model, which acts as a reward signal to learn to choose preferred outputs from GPT-3.

An emerging paradigm for text generation is to prompt (or `ask') LLMs for a desired output \cite{mulla_2023}.
This works by feeding an input prompt or `query' (with a series of examples for a one- or few-shot setting) to a LLM.
This paradigm has inspired a new research direction called {\em prompt engineering}.
One of the most common approaches to prompt engineering involves prepending a string to the context given to a LLM for generation \cite{liu_2023}. For controllable text generation (CTG), such a prefix must contain a control element, such as a keyword that will guide the generation \cite{mulla_2023}. 

Questions are one of the most basic methods used by teachers to educate. As this learning method is so broad, it uses many organizational taxonomies which take different approaches to divide questions into groups.
One popular example is Bloom's taxonomy \cite{krathwohl_2002}, which divides educational material into categories based on student's learning goals.
Another example is a difficulty-level taxonomy, which usually divides questions into 3 categories of easy, medium, and hard \cite{perez_2012}.
By combining CTG and these question taxonomies, we open doors for question generation by prompting LLMs to meet specifications of the educational domain.

\vspace{-2mm} \section{Methodology}\label{methods}
\subsection{Controllable Generation Parameters}
Parameter settings used in this paper were guided by preliminary experimentation. Firstly, `long' context passages (6-9 sentences) empirically appeared to improve generation. Secondly, the few-shot setting outperformed the zero-shot setting, with five-shot (i.e., with 5 context/related question type pairs included in the prompt) performing best. 
Few-shot generation is where prompts consist of an instruction (e.g., "Generate easy questions."), examples (e.g., set of n context/easy question pairs), and the desired task (e.g., context to generate from). 
Thirdly, there was not a large enough sample size to definitively say which question taxonomies are superior to use as control elements for CTG.
Two representative taxonomies were chosen for the experiments in Section \ref{assessment_experiment}: Bloom's taxonomy \cite{krathwohl_2002} (which includes {\em remembering}, {\em understanding}, {\em applying}, {\em analyzing}, {\em evaluating}, and {\em creating} question types) and a difficulty-level taxonomy (which includes {\em beginner}, {\em intermediate}, and {\em advanced} question types) \cite{perez_2012}.
These taxonomies approach the organization of questions in different ways, by the learning goal and by complexity respectively. This creates an interesting comparison among the taxonomic categories to help explore the limits of the CTG approach.

\vspace{-4mm} \subsection{Teacher Assessment Experiment}\label{assessment_experiment}
\vspace{-2mm} \paragraph{Question Generation} The human assessment experiment was conducted with candidates generated in the machine learning (ML) and biology (BIO) domains. There are 68 `long' context passages (6-9 sentences) pulled from Wikipedia (31 are ML, 37 are BIO). Using hand-crafted examples for 5-shot prompting, InstructGPT was prompted to generate 612 candidate questions.\footnote{The passages, few-shot examples, prompt format, taxonomic level definitions, annotator demographics and raw results are available: \url{https://tinyurl.com/y2hy8m4p}.} Each passage has 9 candidates, one with each taxonomic category as the control element.

\vspace{-2mm} \paragraph{Annotators} There are two cohorts of annotators, BIO and ML. The 11 BIO annotators have biology teaching experience at least at a high school level, and were recruited on the freelance platform Up Work. The 8 ML annotators have CS, ML, AI, math or statistics teaching experience at a university level, and were recruited through word of mouth at McGill and Mila.
All of the annotators are proficient English speakers and are from diverse demographics.
Their teaching experience ranges from 1-on-1 tutoring to hosting lectures at a university.
The experiments are identical for both cohorts. As such, the experiment is explained in a domain-agnostic manner. The results will be presented separately, as the goal of this work is not to show identical trends between the two domains, but that CTG is appropriate for education in general.

\vspace{-2mm} \paragraph{Metrics} Each annotator was trained to assess the generated candidates on two of four quality metrics, as well as a \textit{usefulness} metric. This division was done to reduce the cognitive load on an individual annotator.
The quality metrics are: \textit{relevance} (binary variable representing if the question is related to the context),
\textit{adherence} (binary variable representing if the question is an instance of the desired question taxonomy level);
and \textit{grammar} (binary variable representing if the question is grammatically correct),
\textit{answerability} (binary variable representing if there is a text span from the context that is an answer/leads to one).
The \textit{relevance}, \textit{grammar},\footnote{Despite not being a teacher's opinion, this is evaluated because we want to know the model's success here without relying on automatic assessment.} \textit{answerability}, and \textit{adherence} metrics are binary as they are objective measures, often seen in QG literature to assess typical failures of LLMs such as hallucinations or malformed outputs \cite{mulla_2023}. The subjective metric assessed, the \textit{usefulness} metric, is rated on a scale because it is more nuanced. This is defined by a teacher's answer to the question: “Assume you wanted to teach about context X. Do you think candidate Y would be useful in a lesson, homework, quiz, etc.?”
This ordinal metric has the following four categories: \textit{not useful}, \textit{useful with major edits} (taking more than a minute), \textit{useful with minor edits} (taking less than a minute), and \textit{useful with no edits}. If a teacher rates a question as \textit{not useful} or \textit{useful with major edits} we also ask them to select from a list of reasons why (or write their own).

\vspace{-2mm} \paragraph{Reducing Bias} We first conducted a pilot study to ensure the metrics and annotator training were unambiguous. We randomized the order of candidates presented and asked annotators to rate one metric at a time to avoid conflation. We included unmarked questions in order to ascertain if the annotators were paying attention. These questions were obviously wrong (e.g., a random question from a different context, a candidate with injected grammatical errors). Any annotators who did not agree on a minimum of $80\%$  of these `distractor' questions were excluded. The annotators' performance on these is discussed in Section \ref{annotator_agreement}.

\vspace{-2mm} \section{Results and Analysis}\label{results}
\vspace{-2mm} \subsection{Generation Overlap}\label{overlap}
We observed overlaps within the generated candidates for this experiment. Specifically, despite having different control elements, sometimes the LLM  generates the same question for a given context passage twice. 
As a result, out of $612$ candidates, there are $540$ unique ones ($88.24\%$ are unique). We believe this overlap is low enough so the generated candidates are still sufficiently diverse for a teacher's needs. It is important to keep in mind that this overlap is not reflected in the following results, as teachers were asked to rank every candidate independently. Future work by the authors will remove this independence assumption.

\vspace{-4mm} \subsection{Annotator Agreement} \label{annotator_agreement}
All of the participants annotated candidates from 6 context passages. In order to assess their agreement on the task, they annotated a $7^{th}$ passage that was the same for all annotators in a given domain cohort. The results for each metric are reported in Table \ref{mean_and_agreement_metrics}. In both domains, \textit{relevance}, \textit{grammar}, and \textit{answerability} have between $85\%$ and $100\%$ observed agreement. The \textit{adherence} metric has lower agreement, between $60\%$ and $80\%$. Since this metric is more complex than the others and captures the annotators' interpretations of the question taxonomies, we consider this moderate agreement to be acceptable and expected.

Unlike the binary metrics, all candidates were rated on \textit{usefulness} by two annotators. As before, only one context passage, the agreement on which is presented in Table \ref{mean_and_agreement_metrics}, was seen by all annotators. Section \ref{results_useful} discusses the aggregation of the \textit{usefulness} scores on the rest of the data.
In both cohorts, the observed agreement on \textit{usefulness} is around $63\%$. This metric is defined according to a teacher's opinion, and as such is subjective. Thus, the lower agreement between annotators is to be expected. Using Cohen's $\kappa$ to measure the agreement yields a $\kappa = 0.537$ for the ML cohort and a $\kappa = 0.611$ for the BIO cohort, which implies moderate and substantial agreement respectively \cite{cohen_kappa}.
Additionally, the agreement of the annotators on the included `distractor' candidates for this metric (see Section \ref{assessment_experiment}) is $\kappa = 1$ (i.e., perfect agreement), which shows that the annotators agree on the fundamental task but might find different questions useful for their particular approach to teaching.
\begin{table} \begin{center}
    \caption{The quality metrics' mean ($\mu$), standard deviation ($\sigma$), and observed agreement (i.e., \% of the time the annotators chose the same label).} \label{mean_and_agreement_metrics}
    \begin{tabular}{| c || c | c || c | c |}
    \hline
    \textbf{Metric} & $\mu \pm \sigma$ (ML) & Agreement \% (ML) & $\mu \pm \sigma$ (BIO) & Agreement \% (BIO) \\ \hline
    Relevance & 0.967$\pm$0.180 & 100 & 0.972$\pm$0.165 & 100 \\ \hline
    Grammar & 0.957$\pm$0.204 & 92.6 & 0.970$\pm$0.170 & 100 \\ \hline
    Adherence & 0.674$\pm$0.470 & 62.2 & 0.691$\pm$0.463 & 79.9 \\ \hline
    Answerability & 0.914$\pm$0.282 & 94.4 & 0.930$\pm$0.256 & 86.7 \\ \hline
    Usefulness & 3.509$\pm$0.670 & 62.7 & 3.593$\pm$0.682 & 62.8 \\ \hline
    \end{tabular}
\end{center} \vspace{-8mm} \end{table}

\subsection{Quality Metrics}
Three quality metrics, \textit{relevance}, \textit{grammar}, and \textit{answerability}, are consistently high for all generated candidates (see in Table \ref{mean_and_agreement_metrics}).
The fourth quality metric, \textit{adherence}, varies across the taxonomic categories as seen in Figure \ref{adherence}. This variation is similar within the two domains. As might be expected, the more objective categories are easier for the LLM to generate. For instance, looking only at the `remembering' category has an \textit{adherence} of $83.3\%$ for the ML cohort and $91.7\%$ for the BIO cohort. This category is intended to ask for a student to recall a fact or definition. This might be simple for the LLM to replicate by identifying a relevant text span, and reflects the traditional QG task. By contrast, asking a LLM to generate a `creating' question is a more open-ended problem, where a text span from the context may not be the answer. Accordingly, the model struggles on this less constrained task, and has an \textit{adherence} of only $40.0\%$ for the ML cohort and $36.1\%$ for the BIO cohort.

\vspace{-4mm} \subsection{\textit{Usefulness} Metric}\label{results_useful}
\noindent The \textit{usefulness} metric's ordinal categories (see Section \ref{assessment_experiment}) are mapped from $1$ ({\em not useful}) to $4$ ({\em useful with no edits}).
The average usefulness for all candidates is 3.509 for the ML cohort and 3.593 for the BIO cohort. Note that an individual candidate's usefulness is already the average score between two annotator's ratings, and the whole average usefulness is the average across all candidates.
This is a highly promising result showing that on average teachers find that these generated candidates will be useful in a classroom setting.
\begin{figure} \vspace{-4mm}
     \centering
     \begin{subfigure}[b]{0.49\textwidth}
         \centering
         \includegraphics[width=\textwidth]{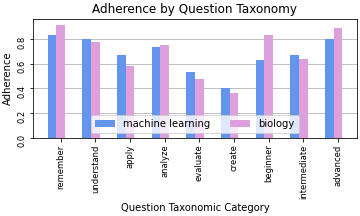}
         \caption{\textit{Adherence} by taxonomic category.}
         \label{adherence}
     \end{subfigure}
     \hfill
     \begin{subfigure}[b]{0.49\textwidth}
         \centering
         \includegraphics[width=\textwidth]{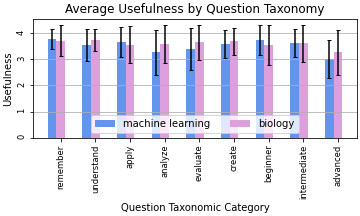}
         \caption{Avg. usefulness by taxonomic category.}
         \label{usefulness}
     \end{subfigure}
        \caption{Visualizations of the \textit{usefulness} and \textit{adherence} metrics.}
\vspace{-4mm} \end{figure}

\noindent There is no significant difference between the usefulness scores of any of the question taxonomy categories, though some variation is present (see Figure \ref{usefulness}). On average, each of the question taxonomies are rated between \textit{useful with minor edits} and \textit{useful with no edits} (i.e., $[3,4]$).
Considering the \textit{adherence} that differs across categories, it is also important to note that a question which does not adhere to its question taxonomy can still be useful in a different way than intended. $56.8\%$ of the time the reason cited for `not useful' candidates is related to their grammar or phrasing. This can possibly be fixed by a filter that removes malformed questions, but it will lower the available diversity of questions.

\vspace{-2mm} \subsubsection{Conclusion}
This work takes steps to demonstrate the realistic usefulness of applying CTG to generate educational questions.
The results show that CTG is a highly promising method that teachers find useful in a classroom setting.
We do not include baselines because the goal is not to show these questions are better than others, only to show they are of high enough quality.
Limitations include the single LLM considered, the independence assumption seen in Section \ref{overlap}, and the lack of comparison between human and machine-authored questions. The authors plan to explore these avenues in future work.
Applying generated candidates to form real-world lessons and evaluate their impact will demonstrate their ultimate value. CTG could pave the way for a new approach to education and transform the experiences of millions of teachers and students.

\paragraph{Acknowledgements} We'd like to thank Mitacs for their grant for this project, and CIFAR for their continued support. We are grateful to both the annotators for their time and the anonymous reviewers for their valuable feedback.

\vspace{-3mm}
\bibliographystyle{splncs04}

\end{document}